\documentclass[a4paper,conference]{IEEEtran}
\IEEEoverridecommandlockouts
\usepackage{cite}
\usepackage{amsmath,amssymb,amsfonts}
\usepackage{algorithmic}
\usepackage{graphicx}
\usepackage{textcomp}
\usepackage{xcolor}
\usepackage{balance}
\def\BibTeX{{\rm B\kern-.05em{\sc i\kern-.025em b}\kern-.08em
    T\kern-.1667em\lower.7ex\hbox{E}\kern-.125emX}}
\begin{document}

\title{Short-Term Load Forecasting Using Time Pooling Deep Recurrent Neural Network}

\author{\IEEEauthorblockN{Elahe Khoshbakhti Vaygan}
\IEEEauthorblockA{\textit{Faculty of ECE} \\
\textit{Qom University of Technology}\\
Qom, Iran \\
khoshbakhti.e@qut.ac.ir}
\and
\IEEEauthorblockN{Roozbeh Rajabi}
\IEEEauthorblockA{\textit{Faculty of ECE} \\
\textit{Qom University of Technology}\\
Qom, Iran \\
rajabi@qut.ac.ir}
\and
\IEEEauthorblockN{Abouzar Estebsari}
\IEEEauthorblockA{\textit{School of the Built Environment and Architecture} \\
\textit{London South Bank University}\\
London, United Kingdom \\
estebsaa@lsbu.ac.uk}
}


\maketitle

\IEEEpubidadjcol

\begin{abstract}
Integration of renewable energy sources and emerging loads like electric vehicles to smart grids brings more uncertainty to the distribution system management. Demand Side Management (DSM) is one of the approaches to reduce the uncertainty. Some applications like Nonintrusive Load Monitoring (NILM) can support DSM, however they require accurate forecasting on high resolution data. This is challenging when it comes to single loads like one residential household due to its high volatility. In this paper, we review some of the existing Deep Learning-based methods and present our solution using Time Pooling Deep Recurrent Neural Network. The proposed method augments data using time pooling strategy and can overcome overfitting problems and model uncertainties of data more efficiently. Simulation and implementation results show that our method outperforms the existing algorithms in terms of RMSE and MAE metrics.
\end{abstract}


\section{Introduction}
One of the main objectives of distribution systems is to ensure continuity of service to final customers and make a balance between supply and demand. Therefore, it is crucial to estimate both generation and load in advance to manage system in operation more efficiently \cite{RN1}. Load forecasting has been a key element in Distribution Management Systems (DMS) in operation and planning analyses as storing electric energy in large amount is very costly. An accurate estimation of load help match supply to avoid imbalances due to power deficit or surplus.

The measurements of electrical load over a long period of time not only show different values of loads over a periodical cycle (e.g. 24 hours), but also realize some trends of load variation in midterm or long-term horizons \cite{RN2}, \cite{RN3}.  Therefore, the load forecasting methods aim at predicting the load values in different time horizons using historic measurements, load trends analysis, experimental rules and mathematical models \cite{RN4}. Some early load forecasting methods included Experimental Smoothing Models, Moving Average, and Auto Regression. The forecasting methods could be classified in three groups based on the techniques used: Gray Prediction Models, Statistical Analysis Models and Non-linear Intelligent Models \cite{RN5}, \cite{RN6}. The most common practical load forecasting methods in traditional passive distribution networks are Auto Regressive Integration Moving Average (ARIMA) and Support Vector Regression (SVR) \cite{RN5} , \cite{RN7}. Integration of more electronic intelligent devices and communication technologies to distribution systems from one side and changing supply paradigms from bulk generation based models to distributed generation create o-called Smart Grids. In Smart Grids, there is a great opportunity for many advanced operation and analysis methods to be applied and utilized. These methods include Artificial Intelligence (AI) application like Machine Learning (ML) techniques and Deep Learning (DL) methods \cite{RN8}, \cite{RN9}, \cite{RN10}, \cite{RN11}.

In This paper, an AI method based on deep learning and pooling strategy techniques is proposed to increase the accuracy of load forecasting of the most volatile loads including single residential households. Similar to supervised multi-layer neural networks, the historic data is divided into two groups containing test and training sets, pooling test and pooling train, respectively. Before presenting our model, the complexity of load forecast in single residential customers and the parameters involved is briefed, and a literature review of ARIMA, SVR, Recurrent Neural Network (RNN) and Deep RNN is presented. The main contribution of this paper is to use time pooling strategy to enhance the performance of deep recurrent neural networks, overcoming the overfitting problem, and better modeling of the data.

The rest of the paper is organized as follows. Section \ref{sec:review} reviews the challenges regarding load forecasting and its applications. In this section previous methods used for short-term load forecasting also summarized. The proposed method using pooling strategy and deep learning is presented in \ref{sec:proposed}. Section \ref{sec:experiments} describes the used dataset and experiments to evaluated the proposed TPRNN. Finally section \ref{sec:conclusion} concludes the paper.

\section{Literature Review}\label{sec:review}
The residential appliances cover a wide range of devices with different characteristics, duty cycles, switching requirements, and power consumption. They are also different in terms of load types, from resistive types to inductive loads. This would make the aggregated profile very volatile with sharp increases or decreases of power in short periods of time. Moreover, the consumption behavior of residential customers is not so predictable and there are always a high level of uncertainty. The uncertainty is due to different cultural and social consumption patterns, weather conditions which could be intermittent, and residual loads considered as noise. Since residential loads have a substantial share in total demand of distribution systems, the prediction of such stochastic loads is essential to manage the system. The importance of more accurate load forecasting is stressed when the system faces intermittent behavior of renewable energy sources (RES) widely integrated into the network. 

\begin{figure}[htbp]
	\centerline{\includegraphics[width=8cm]{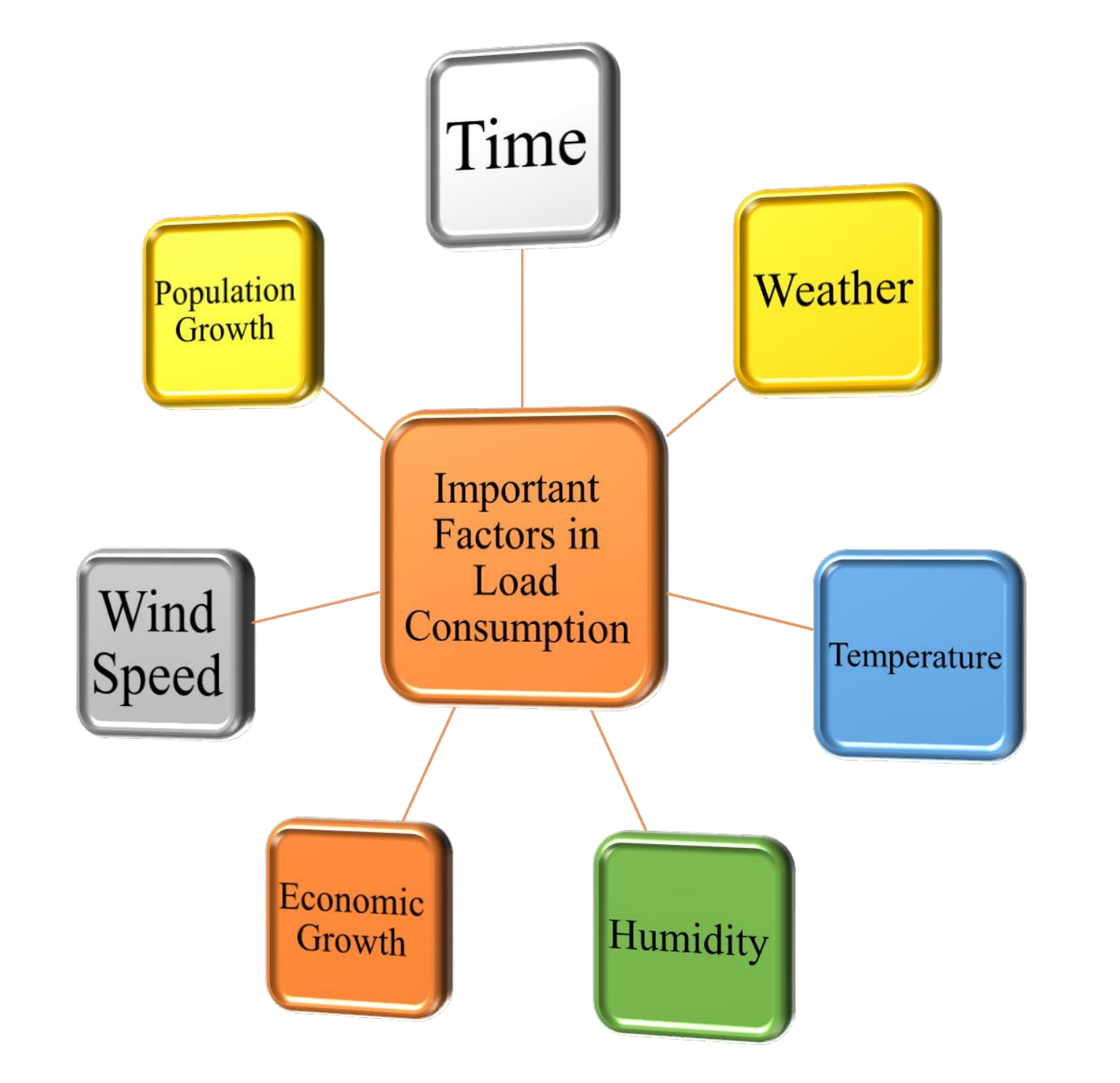}}
	\caption{Important factors having impact on load consumption.}
	\label{fig:Factors}
\end{figure}

Due to high volatility and uncertainty of single residential loads, we believe advanced ML-based methods should be applied to get a high accuracy in the results. DL techniques have been proposed in literature and in several projects as efficient methods, however the accuracy and performance of the applying DLs directly are arguable. In these methods, increase of deep layers is a fundamental approach to improve the accuracy of the results compare to Artificial Neural Networks (ANN), but in single residential load forecasting, the increase of DL layers exponentially increases the intrinsic parameters of the network in the training set. This would eventually grow the system noise which can affect the forecasting results. In summary, it is important to carefully model and adapt the DL network for for particular types of loads and specific forecast time horizons in order to achieve desired accuracy and performance \cite{TSG17_PDRNN}, \cite{RN16}. Some parameters which may be used as extra input to the model include the economical and environmental conditions of the loads such as temperature, humidity, lighting, population growth, wind speed, economical situation, time and policies (Fig. \ref{fig:Factors}).

In the following we briefly review 4 general methods to forecast loads in short-term:   Auto Regressive Integration Moving Average  (ARIMA), Support Vector Regression (SVR), Recurrent Neural Network (RNN) and Deep Recurrent Neural Network (DRNN).

\subsection*{Auto Regressive Integration Moving Average}

Time series models are the most common and popular methods to perform short-term forecasting. The Auto Regressive Integration Moving Average model is derived from Regressive Moving Average method which is time-series based method. In this model, time-series $\mathrm{y}(\mathrm{t})$ is defined based on the previous values $y(t-1), y(t-2), \ldots, y(t-p)$ and stochastic noise. The order of the process depends on the oldest previous value to which $\mathrm{y}(\mathrm{t})$ is returned. This model is formulated as follow:

\begin{equation}
\begin{aligned}
&\mathrm{y}(\mathrm{t})=\delta+\phi_{1} \mathrm{y}(\mathrm{t}+1)+\cdots+ \\
&\phi_{\mathrm{p}} \mathrm{y}(\mathrm{t}-\mathrm{p})+\varepsilon(\mathrm{t})
\end{aligned}
\end{equation}

where $y(t-1), y(t-2), \ldots, y(t-p)$ are the previous values of time series, $\varepsilon(t)$ is the stochastic noise, and $\delta$ is defined as follow \cite{RN5}:

\begin{equation}
\delta \equiv\left(1-\sum_{i=1}^{p} \varphi_{i}\right) \mu
\end{equation}

where $\mu$ is the average value of the process. 

\subsection*{Support Vector Regression}

SVR is based on Vapnik-chervonenkis theory and aim at expanding data sets to those not seen yet \cite{RN7}, \cite{RN17}. SVM can be extended to be used as SVR. For this, a zone named $\varepsilon$ -insensitivities defined around the function as $\varepsilon$ -tube; then the SVR is formulated as follow \cite{RN7}:

\begin{equation}
\begin{gathered}
\mathrm{f}(\mathrm{x})=\left(\begin{array}{c}
\mathrm{x} \\
1
\end{array}\right)\left(\begin{array}{l}
\mathrm{W} \\
\mathrm{b}
\end{array}\right)^{\mathrm{T}}=w^{T}+\mathrm{b} \\
\mathrm{x}, \mathrm{w} \in \mathrm{R}^{\mathrm{m}+1}
\end{gathered}
\end{equation}

\subsection*{Recurrent Neural Network}

The Recurrent Neural Network is based on Long Short-Term Memory (LSTM) methods which is able to exploit the interdependence in time-series over long periods of time. This makes the predictions of electrical loads more accurate. Considering $\mathrm{x}=\left\{\mathrm{x}_{1}, \mathrm{x}_{2}, \ldots, \mathrm{x}_{\mathrm{T}}\right\}$ as input time series, and $\mathrm{h}=\left\{\mathrm{h}_{1}, \mathrm{~h}_{2}, \ldots, \mathrm{h}_{\mathrm{T}}\right\}$ and $\mathrm{y}=\left\{\mathrm{y}_{1}, \mathrm{y}_{2}, \ldots, \mathrm{y}_{\mathrm{T}}\right\}$ as output time series, which are created by updating the states of memory cell, LSTM decides what dat to be removed from the cell gate \cite{RN19}:

\begin{equation}
\begin{aligned}
&\mathrm{f}(\mathrm{x})= \\
&\sigma\left(w_{f x} x_{t}+w_{f h} h_{t-1}+w_{f c} c_{t-1}+b_{f}\right)
\end{aligned}
\end{equation}

In the next step, LSTM stores the new data as new cell state. this would be the input gate layer named $i_{t}$ and formulated as the following:

\begin{equation}
\mathrm{i}_{\mathrm{t}}=\sigma\left(\mathrm{w}_{\mathrm{ix}} \mathrm{x}_{\mathrm{t}}+\mathrm{w}_{\mathrm{ih}} \mathrm{h}_{\mathrm{t}-1}+\mathrm{w}_{\mathrm{ic}} \mathrm{c}_{\mathrm{t}-1}+\mathrm{b}_{\mathrm{i}}\right)
\end{equation}

A new vector is created, $\mathrm{U}_{\mathrm{t}}$, in order to store the new data that are going to be added as the new state to the cell. This vector is formed as the following equation:

\begin{equation}
\mathrm{U}_{\mathrm{t}}=\mathrm{g}\left(\mathrm{w}_{\mathrm{cx}} \mathrm{x}_{\mathrm{t}}+\mathrm{w}_{\mathrm{ch}} \mathrm{h}_{\mathrm{t}-1}+\mathrm{b}_{\mathrm{c}}\right)
\end{equation}

Then, the old cell state $\mathrm{C}_{\mathrm{t}-1}$ is replaced by the new cell state $\mathrm{c}_{\mathrm{t}}$, using $\mathrm{f}_{\mathrm{t}}$ and $\mathrm{U}_{\mathrm{t}}$:

\begin{equation}
c_{\mathrm{t}}=U_{t}{\mathrm{i}}_{\mathrm{t}}+\mathrm{c}_{\mathrm{t}-1} \mathrm{f}_{\mathrm{t}}
\end{equation}

The output of the output gate is represented by the following equation:

\begin{equation}
o_{t}=\sigma\left(w_{0 x} x_{t}+w_{o h} h_{t-1}+w_{0 c} c_{t-1}+b_{0}\right)
\end{equation}

The new data, which is $\mathrm{h}_{\mathrm{t}}$, would be as follow:

\begin{equation}
\mathrm{h}_{\mathrm{t}}=\mathrm{o}_{\mathrm{t}} \mathrm{l}\left(\mathrm{c}_{\mathrm{t}}\right)
\end{equation}

And the final output would equal $\mathrm{y}_{\mathrm{t}}$:

\begin{equation}
y_{t}=k\left(w_{y h} h_{t}+b_{y}\right)
\end{equation}

Where $\mathbf{w}_{\mathrm{ix}}$, $\mathbf{w}_{\mathrm{fx}}$, $\mathbf{w}_{\mathbf{o x}}$, and $\mathbf{w}_{\mathrm{cx}}$ are input matrices; $\mathrm{w}_{\mathrm{ih}}$, $\mathrm{w}_{\mathrm{fh}}$, $\mathrm{w}_{\mathrm{oh}}$, and $\mathrm{w}_{\mathrm{ch}}$ are recurrent weight matrices; and $\mathbf{w}_{\mathrm{yh}}$ is the hidden output weight matrix.  The matrices $\mathbf{w}_{\mathbf{i c}}$, $\mathbf{w}_{\mathrm{fc}}$, and $\mathrm{w}_{\mathrm{oc}}$ are connection weight matrices, and the vectors $\mathrm{b}_{\mathrm{i}}$, $\mathrm{b}_{\mathrm{f}}$, $\mathrm{b}_{\mathrm{o}}$, $\mathrm{b}_{\mathrm{c}}$, and $\mathrm{b}_{\mathrm{y}}$ are based on bias vector \cite{RN19}.

\subsection*{Deep Recurrent Neural Network}

\begin{figure}[htbp]
	\centerline{\includegraphics[width=9cm]{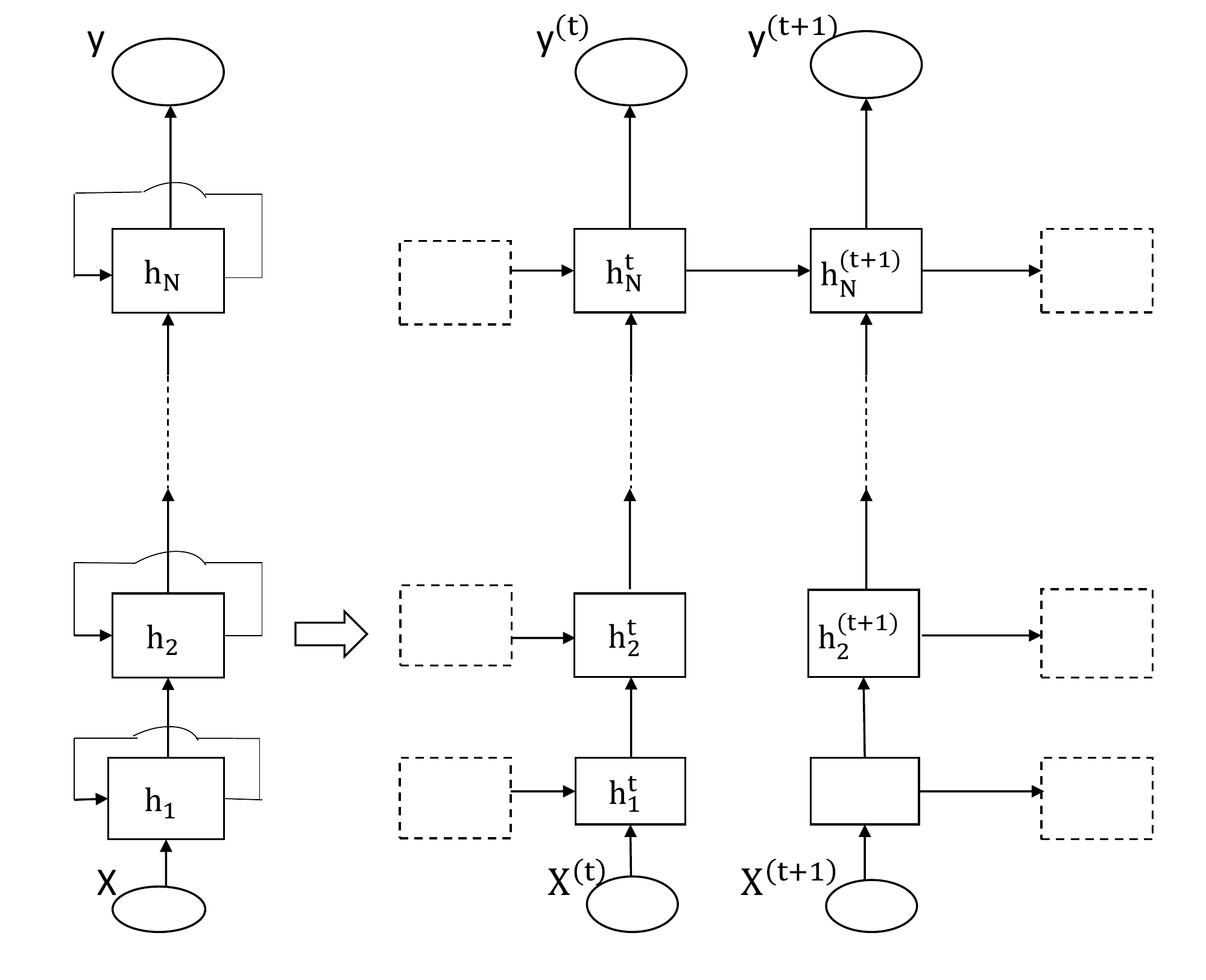}}
	\caption{Structure of DRNN.}
	\label{fig:DRNN_Structure}
\end{figure}

Deep Recurrent Neural Network (DRNN) has more layers in its network compare to RNN and has more benefits comparatively. Figure \ref{fig:DRNN_Structure} shows the structure of DRNN and the functional process of a Deep RNN \cite{TSG17_PDRNN}. For a time-step, $\mathrm{t}$, the parameters of the network neuron on the layer $\mathrm{l}$ updates their values according to the following equations \cite{TSG17_PDRNN}:

\begin{equation}
\mathrm{a}_{l}^{(\mathrm{t})}=\mathrm{b}_{l}+\mathrm{w}_{l} \mathrm{~h}_{l}^{(\mathrm{t}-1)}+\mathrm{U}_{l} \mathrm{x}^{(\mathrm{t})}
\end{equation}

\begin{equation}
\mathrm{h}_{\mathrm{l}}^{(\mathrm{t})}=\text{activation function}\left(\mathrm{a}_{\mathrm{l}}^{(\mathrm{t})}\right){\text { for } \mathrm{l}=1,2, \ldots, \mathrm{N}}
\end{equation}

\begin{equation}
\mathrm{a}_{\mathrm{l}}^{(\mathrm{t})}=\mathrm{b}_{\mathrm{l}}+\mathrm{w}_{\mathrm{l}} \mathrm{h}_{\mathrm{l}}^{(\mathrm{t}-1)}+\mathrm{U}_{l} \mathrm{~h}_{l-1}^{(t)}{\text { for } \mathrm{l}=2,3,...,N}
\end{equation}

\begin{equation}
\mathrm{y}^{(\mathrm{t})}=\mathrm{b}_{\mathrm{N}}+\mathrm{w}_{\mathrm{N}} \mathrm{h}_{\mathrm{N}}^{(\mathrm{t}-1)}+\mathrm{U}_{\mathrm{N}} \mathrm{h}_{\mathrm{N}}^{(\mathrm{t})}
\end{equation}

\begin{equation}
\mathrm{L}=\operatorname{loss} \mathrm{f}\left(\mathrm{y}^{(\mathrm{t})}, \mathrm{y}_{\mathrm{target}}^{(\mathrm{t})}\right)
\end{equation}

where, $\mathbf{x}^{(\mathrm{t})}$ is the input of the time step $t$, $\mathrm{y}^{(\mathrm{t})}$ is the predicted value, and $\mathrm{y}_{\text {target }}^{(\mathrm{t})}$ is the output value. $\mathrm{h}_{\mathrm{l}}^{(\mathrm{t})}$ is the activation function at layer $l$ of the network in time step $t$. $a_{l}^{(t)}$ is the real input value on layer $l$ in time step $t$. The characteristics of the DRNN enables it to learn the uncertainties which are not very precedent.

\section{Proposed Method}\label{sec:proposed}
\begin{figure}[htbp]
	\centerline{\includegraphics[width=5cm]{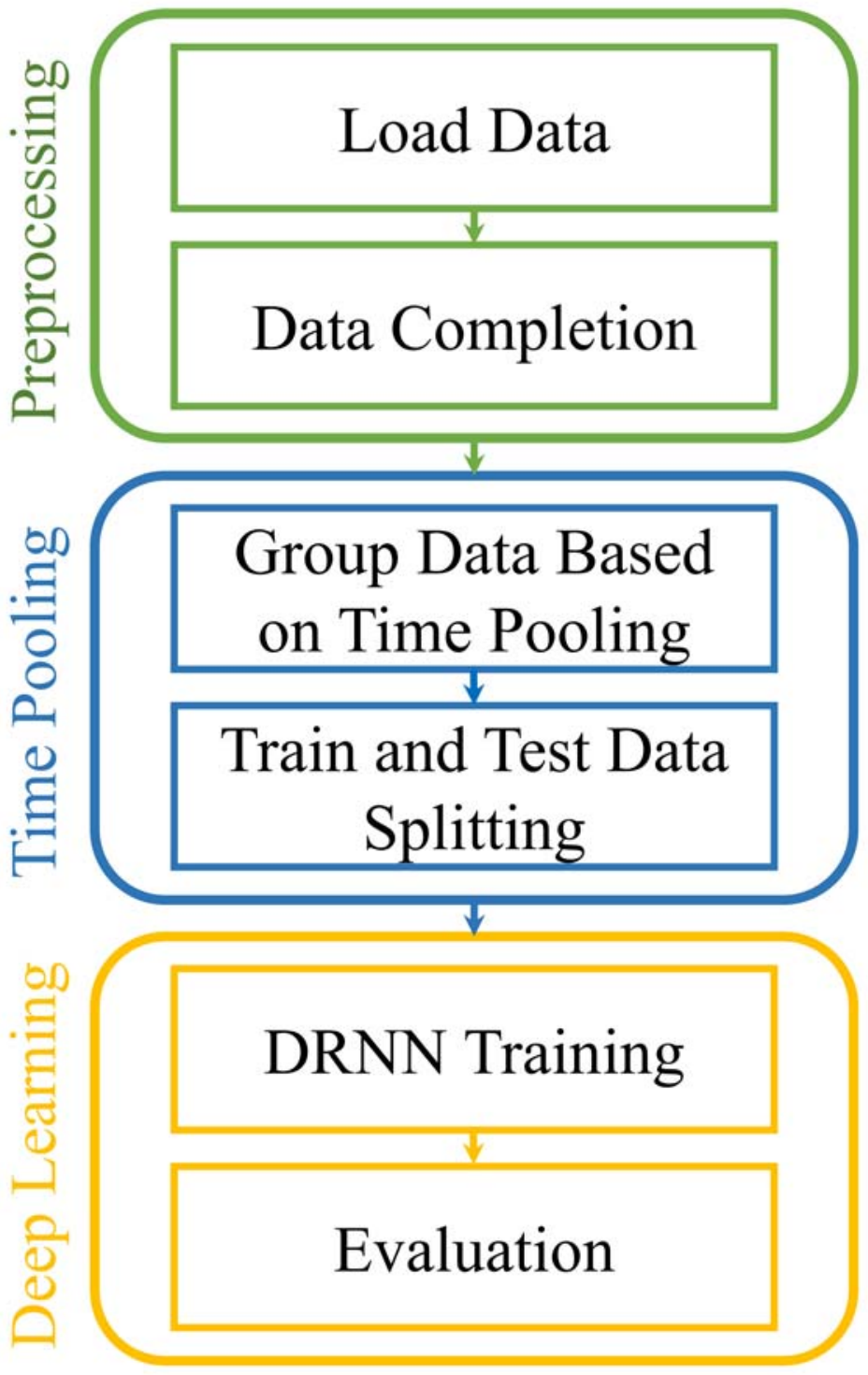}}
	\caption{The proposed time pooling DRNN method.}
	\label{fig:Proposed}
\end{figure}

In this section the proposed methodology is presented. Pooling strategy \cite{TSG17_PDRNN} is used in the proposed method to tackle two main challenges in short-time load forecasting for single house residential applications. Overfitting is one of the major limits in deep learning-based load forecasting. Because of many layers in deep networks, training of these networks needs a large amount of data. Pooling stage can increase the amount of data greatly and avoid overfitting. The second challenge is uncertainty of load consumption data. Due to probabilistic nature of data, it is usually very difficult to model the data. Although some of these uncertainties are due to extrinsic factors such as weather conditions, diversity in training data can improve learning of the uncertainties. So, in the proposed method, pooling strategy is used to augment data and improve learning capabilities of the proposed method. A block diagram of the proposed method is illustrated in Fig. \ref{fig:Proposed}. The first step is loading data and data completion as discussed in \ref{subsec:dataset}. 

After preprocessing stage, the data for each week is partitioned in a time-based manner. Total number of 10,080 measurements are available for each week with the resolution of one minute. This amount of data is divided to half day intervals (N=24*60/2=720) that will generate M number of groups (M=14). Each group will divided into training and test parts generating pools of data to train and test the model. In the next stage, deep recurrent neural network (DRNN) is trained to learn the load consumption profiles and finally the performance of the proposed method is evaluated using test dataset and metrics described in \ref{subsec:results}.

\section{Experiments and Results}\label{sec:experiments}

\subsection{Dataset and Preprocessing}\label{subsec:dataset}
\begin{figure}[htbp]
	\centerline{\includegraphics[width=9cm]{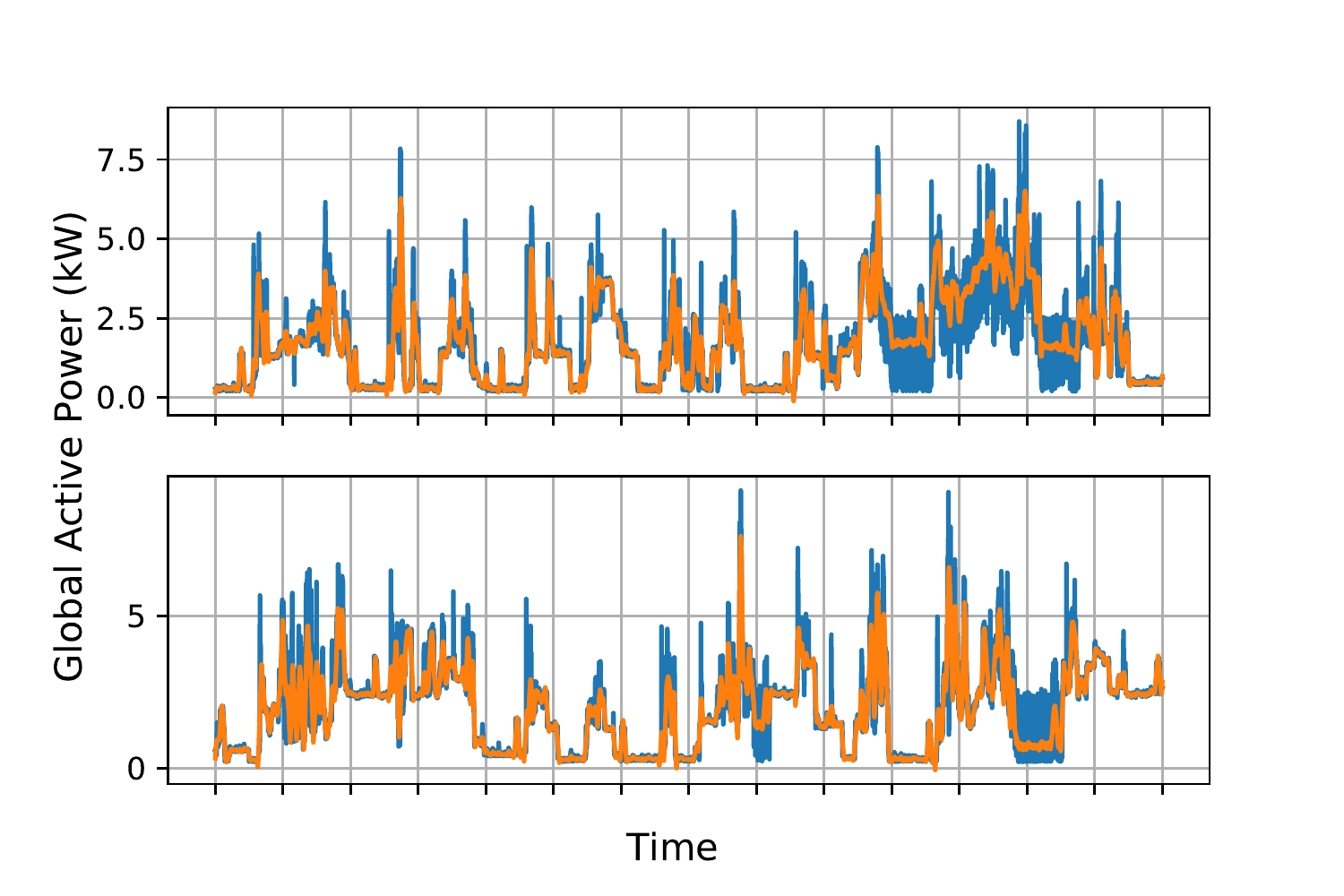}}
	\caption{Sample measurements from the dataset for two consecutive weeks starting from 2006-12-18 and ending to 2006-12-31.}
	\label{fig:Data_TPRNN}
\end{figure}

In this study, the individual household electric power consumption dataset is used to validate the proposed algorithm \cite{Dataset} \footnote{Available online: https://archive.ics.uci.edu/ml/datasets/individual+\newline household+electric+power+consumption}. This dataset consists of 2,075,259 instance points that are measured over a period of almost 4 years. The global active power (GAP) in kW, date and time columns of this dataset are used in this paper. Preprocessing step based on averaging over available data is used to complete missing data. Fig. \ref{fig:Data_TPRNN} shows GAP measurements of two consecutive weeks from the dataset exhibiting roughly similar pattern during a week. This similarity is exploited in the proposed method. 

\subsection{Experiments}
\begin{figure}[htbp]
	\centerline{\includegraphics[width=9cm]{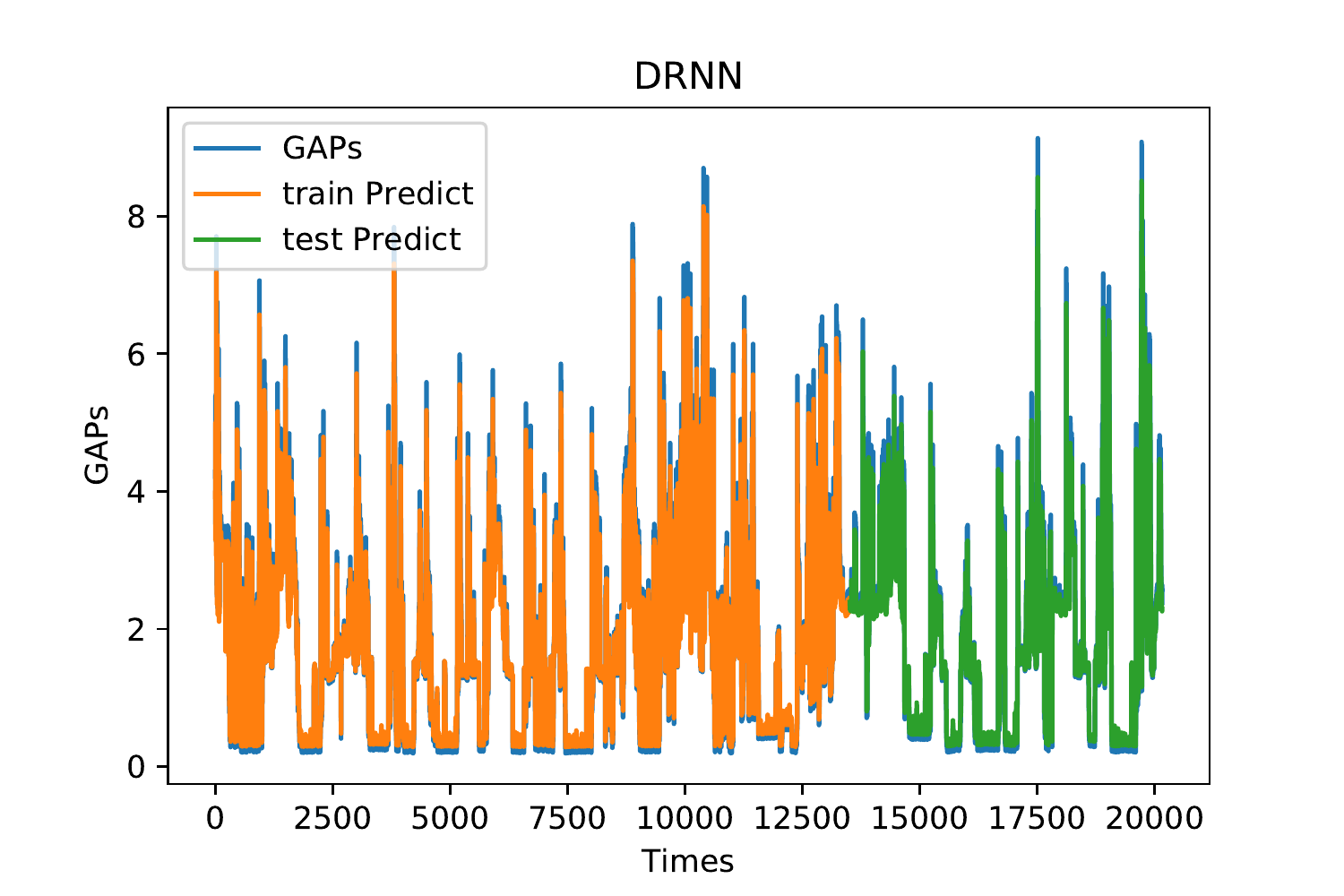}}
	\caption{Results of DRNN method for 2 weeks period.}
	\label{fig:DRNN}
\end{figure}

\begin{figure}[htbp]
	\centerline{\includegraphics[width=9cm]{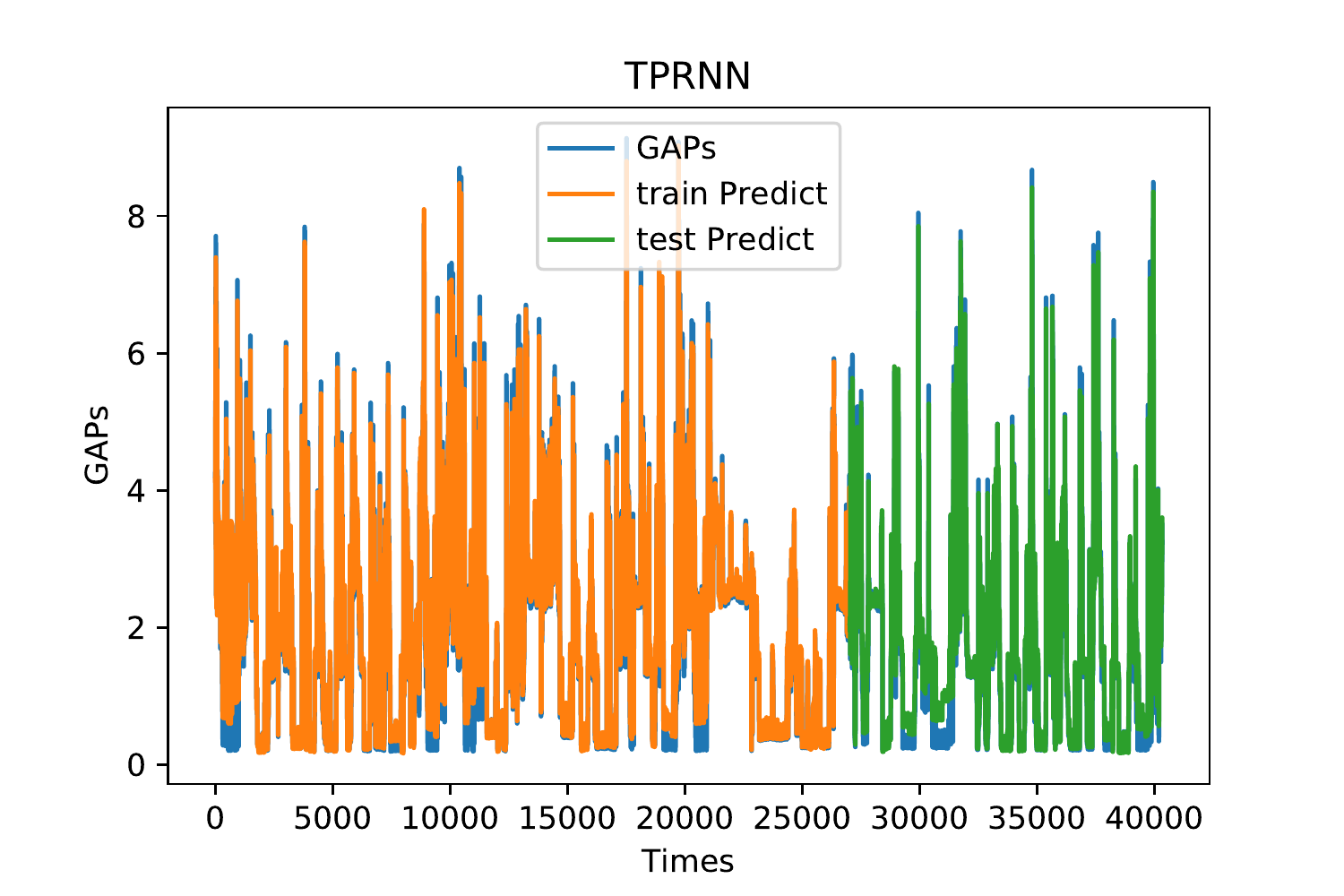}}
	\caption{Results of proposed method (TPRNN) for 4 weeks period.}
	\label{fig:TPRNN}
\end{figure}

All experiments are done using Python programming language on a machine with 8 GB of RAM and Intel core i5-5200U up to 2.7 GHz CPU. The parameters are set as follows: N=720, M=14, training to test ratio: 67/23. 
Fig. \ref{fig:DRNN} and \ref{fig:TPRNN} show the training and test results of two methods DRNN and TPRNN (proposed method) respectively.

\subsection{Results and Discussion}\label{subsec:results}
To evaluate the proposed algorithm three widely used metrics are employed: Root Mean Squared Error (RMSE) and Mean Absolute Error (MAE). These metrics are defined as follows:

\begin{equation}
\mbox{RMSE}=\sqrt{\frac{1}{N}\sum_{i=1}^{N}(y_i-\hat{y_i})^2}
\label{eq:RMSE}
\end{equation}

\begin{equation}
\mbox{MAE}=\frac{1}{N}\sum_{i=1}^{N}|y_i-\hat{y_i}|
\label{eq:MAE}
\end{equation}
where $N$ is the total number of sample points. $y_i$ and $\hat{y_i}$ are the true and estimated values respectively.

Results are compared with different algorithms including SVR, ARIMA, RNN and DRNN in Table. \ref{tbl:comparison}. The proposed method (TPRNN) exhibits better performance in comparison with other state-of-the-art methods.

\begin{table}[htbp]
	\caption{Comparison Results Based on RMSE and MAE.}
	\begin{center}
		\begin{tabular}{|c|c|c|c|}
			\hline
			\textbf{Method}&{RMSE}&{MAE} \\
			\hline
			\textbf{SVR} &0.96& 0.77\\
			\hline
			\textbf{ARIMA} & 0.81& 0.75\\
			\hline
			\textbf{RNN} & 0.75& 0.55\\
			\hline
			\textbf{DRNN} & 0.39& 0.20\\
			\hline
			\textbf{TPRNN} & 0.37& 0.19\\
			\hline
		\end{tabular}
		\label{tbl:comparison}
	\end{center}
\end{table}

\section{Conclusion}\label{sec:conclusion}
Short-term load forecasting is an input or prerequisite for many applications in power systems, specially in smart grids. Depending on the application, different types of data are provided to run load forecasting algorithms. The accuracy of the forecasting is important for different applications too. In this paper, we presented a new short-term load forecasting method based on time pooling DRNN to predict the power demand of single residential customers for applications which require very short-term forecast. An application which could use this method is nonintrusive load monitoring (NILM). A high resolution historic data is used for training the model.

Load profiles demonstrate similarities in time. Based on these similarities, in this paper time-pooling strategy in conjunction with deep recurrent neural networks is used to solve the short-term load forecasting problem. Using pooling strategy has two fold advantages. Firstly it satisfies the need for large amount of data in deep learning methods and helps to model uncertainties in the dataset precisely. Results demonstrates that the proposed method can effectively forecast load consumption profile. Further research could be done including more precise time pooling using clustering algorithms to put similar power consumption patterns in a group.

\balance

\bibliographystyle{IEEEtran}
\bibliography{IEEEabrv,refs}

\begin{thebibliography}{10}
\providecommand{\url}[1]{#1}
\csname url@samestyle\endcsname
\providecommand{\newblock}{\relax}
\providecommand{\bibinfo}[2]{#2}
\providecommand{\BIBentrySTDinterwordspacing}{\spaceskip=0pt\relax}
\providecommand{\BIBentryALTinterwordstretchfactor}{4}
\providecommand{\BIBentryALTinterwordspacing}{\spaceskip=\fontdimen2\font plus
\BIBentryALTinterwordstretchfactor\fontdimen3\font minus
  \fontdimen4\font\relax}
\providecommand{\BIBforeignlanguage}[2]{{%
\expandafter\ifx\csname l@#1\endcsname\relax
\typeout{** WARNING: IEEEtran.bst: No hyphenation pattern has been}%
\typeout{** loaded for the language `#1'. Using the pattern for}%
\typeout{** the default language instead.}%
\else
\language=\csname l@#1\endcsname
\fi
#2}}
\providecommand{\BIBdecl}{\relax}
\BIBdecl

\bibitem{RN1}
A.~Garulli, S.~Paoletti, and A.~Vicino, ``Models and techniques for electric
  load forecasting in the presence of demand response,'' \emph{IEEE
  Transactions on Control Systems Technology}, vol.~23, no.~3, pp. 1087--1097,
  2014.

\bibitem{RN2}
A.~Estebsari and R.~Rajabi, ``Single residential load forecasting using deep
  learning and image encoding techniques,'' \emph{Electronics}, vol.~9, no.~1,
  p.~68, 2020.

\bibitem{RN3}
R.~Rajabi and A.~Estebsari, ``Deep learning based forecasting of individual
  residential loads using recurrence plots,'' in \emph{2019 IEEE Milan
  PowerTech}.\hskip 1em plus 0.5em minus 0.4em\relax IEEE, Conference
  Proceedings, pp. 1--5.

\bibitem{RN4}
H.~N. Rafsanjani, ``Factors influencing the energy consumption of residential
  buildings: a review,'' in \emph{Construction Research Congress 2016},
  Conference Proceedings, pp. 1133--1142.

\bibitem{RN5}
F.~Mahia, A.~R. Dey, M.~A. Masud, and M.~S. Mahmud, ``Forecasting electricity
  consumption using {ARIMA} model,'' in \emph{2019 International Conference on
  Sustainable Technologies for Industry 4.0 (STI)}.\hskip 1em plus 0.5em minus
  0.4em\relax IEEE, Conference Proceedings, pp. 1--6.

\bibitem{RN6}
A.~Srivastava, A.~S. Pandey, and D.~Singh, ``Notice of violation of ieee
  publication principles: Short-term load forecasting methods: A review,'' in
  \emph{2016 International conference on emerging trends in electrical
  electronics \& sustainable energy systems (ICETEESES)}.\hskip 1em plus 0.5em
  minus 0.4em\relax IEEE, Conference Proceedings, pp. 130--138.

\bibitem{RN7}
M.~Awad and R.~Khanna, \emph{Support vector regression}.\hskip 1em plus 0.5em
  minus 0.4em\relax Springer, 2015, pp. 67--80.

\bibitem{RN8}
T.~Hossen, S.~J. Plathottam, R.~K. Angamuthu, P.~Ranganathan, and H.~Salehfar,
  ``Short-term load forecasting using deep neural networks ({DNN}),'' in
  \emph{2017 North American Power Symposium (NAPS)}.\hskip 1em plus 0.5em minus
  0.4em\relax IEEE, Conference Proceedings, pp. 1--6.

\bibitem{RN9}
A.~Almalaq and G.~Edwards, ``A review of deep learning methods applied on load
  forecasting,'' in \emph{2017 16th IEEE international conference on machine
  learning and applications (ICMLA)}.\hskip 1em plus 0.5em minus 0.4em\relax
  IEEE, Conference Proceedings, pp. 511--516.

\bibitem{RN10}
E.~Busseti, I.~Osband, and S.~Wong, ``Deep learning for time series modeling,''
  \emph{Technical report, Stanford University}, pp. 1--5, 2012.

\bibitem{RN11}
H.~Chen, C.~A. Canizares, and A.~Singh, ``{ANN}-based short-term load
  forecasting in electricity markets,'' in \emph{2001 IEEE power engineering
  society winter meeting. Conference proceedings (Cat. No. 01CH37194)},
  vol.~2.\hskip 1em plus 0.5em minus 0.4em\relax IEEE, Conference Proceedings,
  pp. 411--415.

\bibitem{TSG17_PDRNN}
H.~Shi, M.~Xu, and R.~Li, ``Deep learning for household load forecasting—a
  novel pooling deep rnn,'' \emph{IEEE Transactions on Smart Grid}, vol.~9,
  no.~5, pp. 5271--5280, 2017.

\bibitem{RN16}
X.~Sun, P.~B. Luh, K.~W. Cheung, W.~Guan, L.~D. Michel, S.~S. Venkata, and
  M.~T. Miller, ``An efficient approach to short-term load forecasting at the
  distribution level,'' \emph{IEEE Transactions on Power Systems}, vol.~31,
  no.~4, pp. 2526--2537, 2016.

\bibitem{RN17}
X.~Sun, P.~B. Luh, K.~W. Cheung, W.~Guan, L.~D. Michel, S.~Venkata, and M.~T.
  Miller, ``An efficient approach to short-term load forecasting at the
  distribution level,'' \emph{IEEE Transactions on Power Systems}, vol.~31,
  no.~4, pp. 2526--2537, 2015.

\bibitem{RN19}
J.~Zheng, C.~Xu, Z.~Zhang, and X.~Li, ``Electric load forecasting in smart
  grids using long-short-term-memory based recurrent neural network,'' in
  \emph{2017 51st Annual Conference on Information Sciences and Systems
  (CISS)}.\hskip 1em plus 0.5em minus 0.4em\relax IEEE, Conference Proceedings,
  pp. 1--6.

\bibitem{Dataset}
G.~Hebrail and A.~Berard, ``Individual household electric power consumption
  data set,'' \emph{{\'E}. d. France, Ed., ed: UCI Machine Learning
  Repository}, 2012.

\end{thebibliography}

\end{document}